\pdfoutput=1
\documentclass[11pt]{article}

\usepackage{ACL2023}

\usepackage{times}
\usepackage{latexsym}

\usepackage[T1]{fontenc}
\usepackage[utf8]{inputenc}

\usepackage{microtype}

\usepackage{inconsolata}
\usepackage{graphicx}
\usepackage{hyperref}
\usepackage{xspace}
\usepackage{booktabs}
\usepackage{cleveref}

\newcommand{\modelname}{\texttt{RASO}\xspace}

\usepackage{multirow}
\title{
Generate then Select: Open-ended Visual Question Answering Guided by World Knowledge
}

\author{Xingyu Fu\textsuperscript{\rm 1}\thanks{\ \ Work done during internship at AWS AI Labs} \hspace{0.1em},  
Sheng Zhang\textsuperscript{\rm 2},  
Gukyeong Kwon\textsuperscript{\rm 2}, 
Pramuditha Perera\textsuperscript{\rm 2},  \\
\textbf{Henghui Zhu\textsuperscript{\rm 2},
Yuhao Zhang\textsuperscript{\rm 2},
Alexander Hanbo Li\textsuperscript{\rm 2},
William Wang\textsuperscript{\rm 2}, } \\ 
\textbf{Zhiguo Wang\textsuperscript{\rm 2}, 
Vittorio Castelli\textsuperscript{\rm 2}, 
Patrick Ng\textsuperscript{\rm 2},
Dan Roth\textsuperscript{\rm 2}, 
Bing Xiang\textsuperscript{\rm 2}} \\
\textsuperscript{\rm 1} University of Pennsylvania, \textsuperscript{\rm 2} AWS AI Labs\\
\texttt{xingyuf2@seas.upenn.edu
}}

\begin{document}
\maketitle
\begin{abstract}
% \ww{It would be useful to give more motivation. For example, why the generate-then-select strategy is needed? Your approach is a very nice way to expand the knowledge coverage from in-domain training data. The Self-Consistency (SC) approach uses multiple generations to the ensemble to check consistency in generated results. In contrast, you are using in-domain training data to somehow check the consistency and coverage of the knowledge and expand your world knowledge. Also, I think this motivation is more interesting than the pipeline itself.}
The open-ended Visual Question Answering (VQA) task requires AI models to jointly reason over visual and natural language inputs using world knowledge.
Recently, pre-trained Language Models (PLM) such as GPT-3 have been applied to the task and shown to be powerful world knowledge sources. However, these methods suffer from low knowledge coverage caused by PLM bias -- the tendency to generate certain tokens over other tokens regardless of prompt changes, and high dependency on the PLM quality -- only models using GPT-3 can achieve the best result. 

To address the aforementioned challenges, we propose \modelname: a new VQA pipeline that deploys a gene\underline{ra}te-then-\underline{s}elect strategy guided by w\underline{o}rld knowledge for the first time. 
Rather than following the de facto standard to train a multi-modal model that directly generates the VQA answer, \modelname first adopts PLM to generate all the possible answers, and then trains a lightweight answer selection model for the correct answer.
As proved in our analysis, \modelname expands the knowledge coverage from in-domain training data by a large margin.
% As the first step, it feeds the image information and question into a frozen PLM to retrieve all the possible answers. Then, it trains a light-weight answer selection model for the correct answer. 
We provide extensive experimentation and show the effectiveness of our pipeline by advancing the state-of-the-art by +4.1\% on OK-VQA, without additional computation cost.
Code and models are released at \url{http://cogcomp.org/page/publication_view/1010}

\end{abstract}
\section{Introduction}
\begin{figure}[t]
  \centering
  \includegraphics[width=0.5\textwidth]{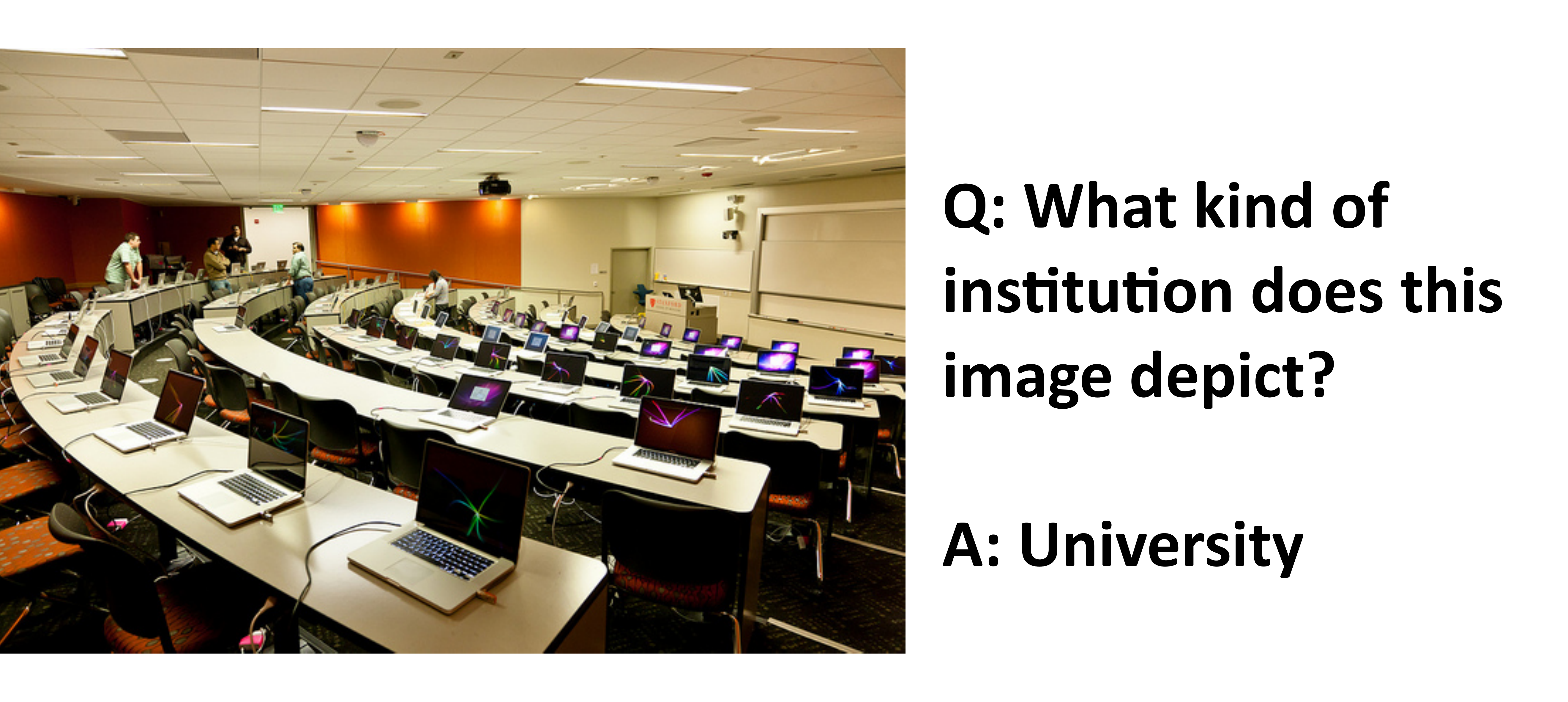}
  \caption{An example data from the OK-VQA dataset, which requires external knowledge not present in the image to answer the question.}
  \label{fig:example}
  % \vspace{-em}
\end{figure}

% \fxy{explainability}
% Recent years have featured a trend toward multimodal research studies \cite{Biten_2022_CVPR,fu-etal-2022-theres}, especially between vision and language（VL).
Open-ended Visual Question Answering (VQA), that requires answering a question based on an image, has received much attention in machine learning research in the past decade \cite{antol2015vqa,balanced_vqa_v2}. Knowledge-based VQA\cite{marino2019ok,AOKVQA} is a variant of VQA, where models have to use external knowledge that is not present in the image to generate the answer.  It is a more challenging problem as it requires joint reasoning over visual and natural language inputs using world knowledge. For example, in Figure \ref{fig:example}, the VQA model needs to conduct multiple levels of inference: to detect the objects in the image (e.g. laptops, whiteboard, etc), to retrieve external world knowledge (e.g, university is an institution and has lecture rooms, lecture rooms have laptops, stairs, and whiteboard, etc), and combine the important visual parts with retrieved knowledge to induce the final answer (e.g. university). 

% The difficulties mainly lie in two parts: how to jointly reason with the given question image pair and retrieve required world knowledge, and how to conduct high-quality open-ended answer generation. 
% In this paper, we specifically focus on the latter problem.

% \begin{figure*}[t]
% \begin{center}
%     \hspace*{-0.25cm}
%     \includegraphics[width=1\linewidth]{figures/main.pdf}
%     % \vspace{-5em}
%     \caption{The main idea of this paper. To overcome the disadvantages of the two kinds of previous methods: limited answer vocabulary and PLM restrictions, we propose to combine these two methods together by having PLMs generate a limited answer vocabulary. Note that in traditional VQA methods, the purple block of ``answer vocab'' is either provided or automatically collected from the training set by frequency. The robots stand for VQA answer selection models. \fxy{remove?}}
%     \label{fig:main}
%     \vspace{-1em}
% \end{center}
% \end{figure*}

\begin{figure*}[ht]
\begin{center}
    \hspace*{-0.25cm}
    \includegraphics[width=1\linewidth]{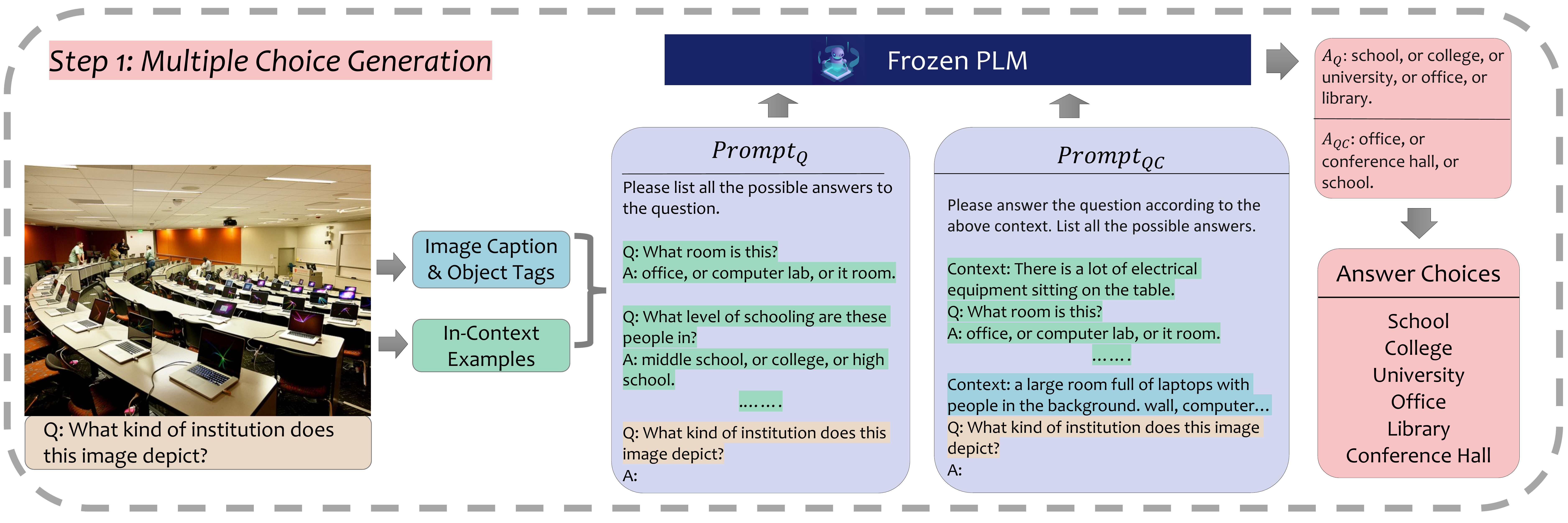}
    % \vspace{-5em}
    \caption{Our multiple choice generation step. Given an image, we use existing tools to get the caption and object tags. We then select most similar examples from the training data and construct the two prompts. We combine the PLM outputs and get the answer choice list. Note that the list is ranked by PLM probability from high to low. More details can be found in Section \ref{sec:method:gen}.}
    \label{fig:gen}
    % \vspace{-1em}
\end{center}
\begin{center}
    \hspace*{-0.5cm}
    \includegraphics[width=1.05\linewidth]{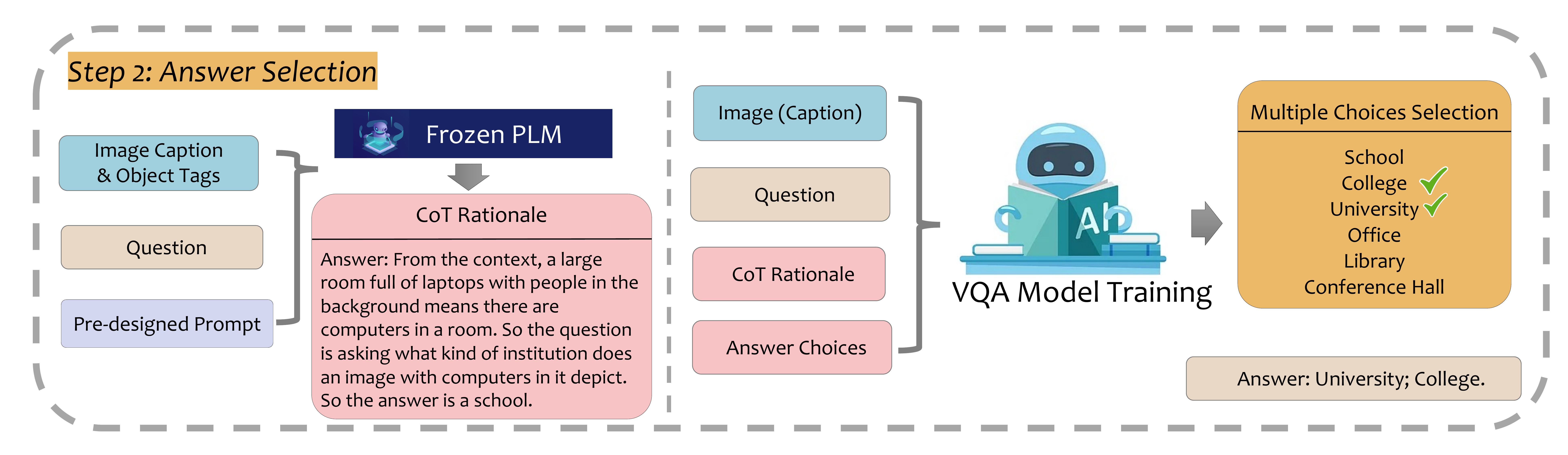}
    % \vspace{-5em}
    \caption{Our answer selection step. Before selecting the final answer, we first use the same PLM to generate a chain-of-thought rationale to guide the process. Then input being the image or its caption, the question, CoT rationale, and answer choices from Step 1, we train a model to output the correct answer. See Section \ref{sec:exp:select} for details about the answer selection models we experiment with.}
    \label{fig:select}
    \vspace{-1em}
\end{center}
\end{figure*}

% As for the first difficulty, several recent papers \cite{marino2021krisp,yang2021empirical,gui-etal-2022-kat} have proved that pre-trained language models such as GPT-3 can be a great source for knowledge retrieval. And this can be combined with previous methods \cite{wu2022multi} that depend on knowledge graphs.
% Although the task is open-ended, traditional VQA methods \cite{shih2016look,garderes-etal-2020-conceptbert,wu2022multi} mainly tackle the problem in the multiple-choice setting -- either with provided multiple choices or with automatically collected answer vocabulary. \GK{Need references}
% In Figure \ref{fig:main} (a), the answer vocababulary is commonly collected from the dataset's training set by frequency, with around two thousands in size \cite{ben2017mutan,marino2021krisp}.
% Regardless that some papers use knowledge graphs for world knowledge retrieval, 
% The major strategy of these methods is to train a classification model over the answer vocab.
% In other words, if the gold answer is outside of the answer vocab, these models will have 0\% accuracy.
% In other words, if the gold answer is outside of the answer vocab, these models will have 0\% accuracy.

% \GK{We are mixing two ideas here (classificaiton-based VQA vs. Generation -based VQA) and (using no world knowledge vs. using world knowledge). If you want to talk about PLM in the next paragraph, the previous paragraph should highlight the limitation of not using world knowledge. Not classification-based VQA.}
In this paper, we focus on improving the important step of external knowledge retrieval.
A common procedure of previous VQA methods \cite{marino2021krisp,wu2022multi} is to retrieve with knowledge graphs from diverse knowledge bases (e.g. Wikipedia \cite{wiki}, ConceptNet \cite{liu2004conceptnet}, etc.), with the results being input to an answer generation model. However, the retrieved knowledge could be noisy, irrelevant, and redundant, and therefore lead to mismatches that limit the VQA performance.
Motivated by the development of large-scale PLMs such as GPT-3 \cite{brown2020language} that obtain state-of-the-art (SOTA) performance in most NLP tasks including text generation \cite{chowdhery2022palm}, more recent approaches PiCA \cite{yang2021empirical} and KAT \cite{gui-etal-2022-kat} propose to retrieve from GPT-3 and achieve better performance for their neat and high-quality knowledge.
Specifically, PiCA directly treats GPT-3 output as the VQA answer, while KAT further uses GPT-3 outputs to train an answer generation model.

\begin{table}[h]
% \small
\centering
\setlength{\tabcolsep}{1.5pt}
\begin{tabular}{c|ccccc}
\hline
 & GPT-J  & UL2 & GPT-3  & OPT & Codex\\ \hline
$Prompt_Q$  & 32.4  & 32.6& -& 34.21& 44.8    \\
$Prompt_{QC}$  & 37.1  &37.5&48.0&37.8&52.9 \\
\bottomrule
\end{tabular}
\caption{Knowledge coverage (\%) of different five PLMs, evaluated on OK-VQA. $Prompt_Q$ means that the prompt to PLM is constructed by the VQA question only, and $Prompt_{QC}$ means that the prompt is constructed by the VQA image and question together. Note that the GPT-3 score is taken from \cite{yang2021empirical}.}
\label{tbl:gen}
\vspace{0.5em}
\end{table}

While achieving SOTA at the time, the two models suffer from the low knowledge coverage caused by PLM bias -- the tendency to generate certain tokens over other tokens despite the prompt changes, and their performance are highly dependent on the PLM quality -- only GPT-3 and Codex can achieve good results. 
As illustrated in Table \ref{tbl:gen}, we report the knowledge coverage percentage of different PLMs on OK-VQA \cite{marino2019ok}, a knowledge-based open VQA dataset. 
We use the accuracy of PiCA as a representation of knowledge coverage, and the first column indicates the PLM input prompts, where $Prompt_Q$ is constructed by VQA question only, and $Prompt_{QC}$ is constructed by image and question together. The top row lists five selected PLMs with parameter size varying from 6.7B to 175B: GPT J \cite{gpt-j}, UL2 \cite{tay2022unifying}, OPT-175B \cite{zhang2022opt}, GPT-3, and Codex \cite{chen2021evaluating}.
Table \ref{tbl:gen} proves that existing VQA approaches using PLMs can only cover less than half (37\% - 53\%) of the required external knowledge.
Further, the small difference (5\% - 8\%) between $Prompt_Q$ and $Prompt_{QC}$ coverage percentages show that PLM bias -- the tendency to generate certain tokens over others given the same question -- is not alleviated by prompt changes such as the inclusion of the image information or not.

To address these challenges, we propose \modelname, a new VQA pipeline that expands world knowledge retrieval by requesting PLMs to generate multiple answer choices, followed by an answer selection model. As shown in Figure \ref{fig:gen}, we first propose a new prompting method to retrieve a long list of possible answers using in-context examples from in-domain training data. 
Note that for the example data in Figure \ref{fig:example}, the PiCA end-task output would be ``office'' as in $A_{QC}$ in Figure \ref{fig:gen}.
With this prompting method, we expand the external knowledge coverage by more than +20\% for each PLM, without additional training data.
Then, as illustrated in Figure \ref{fig:select}, we propose a chain-of-thought (CoT) \cite{wei2022chain} guided answer selection approach. 
% We experiment with four existing VQA methods \cite{gui-etal-2022-kat,khashabi2022unifiedqa,mokady2021clipcap,yang2021empirical} plugged in as the answer selector model and show their results.
By plugging in the previous SOTA method KAT \cite{gui-etal-2022-kat} as the answer selector, we achieve the new SOTA performance 58.5\% (+4.1\%) on the OK-VQA dataset without additional computation effort.

Extensive experiments in Section \ref{sec:exp} suggest that \modelname provides a general way to increase the retrieved world knowledge coverage using PLMs, boosting end-task performance without additional computation cost. We believe our proposed pipeline motivates a new type of generate-then-select VQA method and facilitates future work.

Our main contributions are: (a) We provide a new prompting method using PLMs that extends the retrieved external knowledge coverage by 20\% over previous approaches in VQA; (b) We are the first to propose a general generate-then-select VQA pipeline, different from the de facto tradition of direct generation approaches; (c) We achieve the new SOTA on the challenging OK-VQA benchmark.

\section{Related Work}
\subsection{VQA Methods}
Visual question answering (VQA) has always been one of the most popular topics in the natural language and computer vision community over recent years.
While the VQA task is free-form and open-ended as first proposed in \cite{antol2015vqa}, a large portion of previous methods \cite{shih2016look,anderson2018bottom,lu2019vilbert,garderes-etal-2020-conceptbert} cast it as a classification problem.
It's a common strategy for them to construct a target vocabulary from the dataset's training set by answer frequency, resulting in around two to four thousand candidates in the target vocabulary \cite{ben2017mutan,yu2019deep,marino2021krisp,wu2022multi}.
These methods suffer from the limited answer vocabulary -- if the gold answer is outside of the vocabulary, then there is no way for these models to have the correct answer.

Rather than closed-set classification, several recent methods focus on direct generating for the correct answer \cite{gui-etal-2022-kat,salaberria2023image} using transformer-based models such as T5 \cite{raffel2020exploring}.
Large-scale multi-modal models trained on multiple vision language tasks \cite{alayrac2022flamingo,chen2022pali} have also become popular and achieved good performance on the OK-VQA dataset. However, these models are not publicly available and necessitate a vast quantity of data and computation resources.

Different from all the previous approaches that are either classification or direct generation, our proposed pipeline \modelname is the first approach ever to follow a generate-then-select strategy, as far as this paper is written. We hope to benefit from less computation cost in the selection part compared to direct generation, while keeping the free-form open-ended answer vocabulary from the answer generation part.

% % mostly GPT-3 to benefit from its world knowledge retrieval and reasoning capabilities. 
% In Figure \ref{fig:main} (b), these methods construct a prompt with few-shot VQA examples, and then feed the prompt into PLM for the answer.
% The prompt can be either discrete by turning an image into a caption, or continuous by turning an image into a multi-modal embedding \cite{radford2021learning}. \GK{Not sure mentioning discrete or continuous prompts are necessary at this point. If you want to use these terms, please define or cite references accrodingly.}
% These methods largely depend on the PLM along with prompt-tuning qualilty, while requiring a large amount of training data and high computation resource, and suffering from bias in PLMs. 

\subsection{Knowledge-based VQA}
While significant progress \cite{lu2016hierarchical,anderson2018bottom,lu2019vilbert,jiang2020defense,marino2021krisp,Biten_2022_CVPR} has been made on the most famous VQA benchmarks \cite{antol2015vqa,balanced_vqa_v2,wang2017fvqa,singh2019towards}, researchers start to raise more challenging questions that require external knowledge not inside the image to answer \cite{marino2019ok,zellers2019vcr,park2020visualcomet,AOKVQA,fu-etal-2022-theres}.

Two-step approaches 
\cite{marino2021krisp,wu2022multi,gui-etal-2022-kat,lin-byrne-2022-retrieval,Gap2022,hu2022promptcap,lin2022revive} that explicitly retrieve world knowledge as input to the end-task model have received much attention. However, these methods could retrieve noisy and redundant information that limits the VQA performance, or have low knowledge coverage. In contrast, without retrieving documents, they may suffer from PLM hallucinations. To address these problems, we treat LLM as a world knowledge source with wide coverage, and propose new prompt-engineering methods to retrieve succinct but higher-quality knowledge, represented as answer choices.
\section{Method}

\begin{figure*}[t]
\centering
    \vspace{-0.5em}
\includegraphics[width=\textwidth]{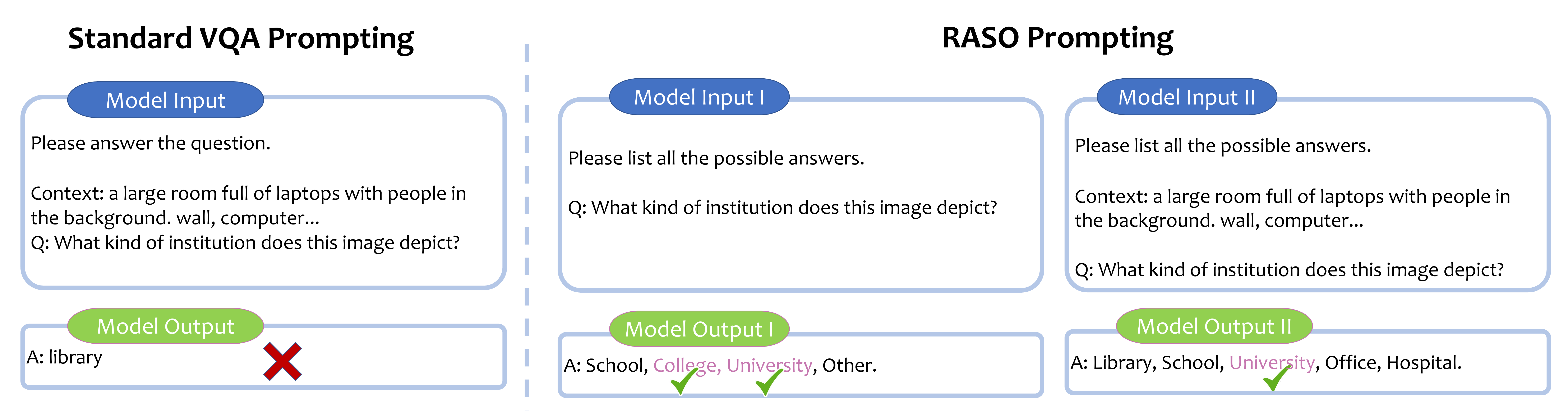}
\caption{An illustration of our proposed prompting method for choice generation enabling larger knowledge retrieval coverage, compared with standard prompting as in PiCA \cite{yang2021empirical}. 
Note that Model Input I and II corresponds to $Prompt_Q$, $Prompt_{QC}$ respectively, and correct answers are highlighted.}
\label{fig:gen_prompt}
\end{figure*}

% \subsection{Overview}
Our method consists of two steps: answer choices generation and answer selection. 
The overview of the proposed model is shown in \Cref{fig:gen,fig:select}. 
\textbf{Problem Formulation}
Given a training dataset $D=\left\{\left(v_i, q_i, a_i\right)\right\}_{i=1}^N
$, where $v_i$ denotes the i-th training image and $N$ is the total number of the training images, $q_i$ and $a_i$ represent the i-th question and its corresponding answer, respectively.
We deploy a generate-then-select strategy to first generate a set of answer choices using a frozen PLM $g$, then trains a model $p$ to select the correct answer from it.
$g$ takes $v_i$ and $q_i$ as inputs, and generates all the possible answers $\hat{A_i} =\{\hat{a_{i0}}, \hat{a_{i1}}, \hat{a_{i2}}, ...\}$.
Finally, $p$ takes $v_i$, $q_i$, and $\hat{A_i}$ as inputs and learns a set of parameters $\theta$ to select from $\hat{A_i}$ for the final answer.

\subsection{Answer Choices Generation}
\label{sec:method:gen}
We design our generation process with inspirations from the previous work \cite{yang2021empirical,gui-etal-2022-kat}. As demonstrated in \Cref{fig:gen,fig:gen_prompt}, we follow a similar strategy to use few-shot in-context learning and leverage a frozen PLM $g$ to generate all the possible answer choices.

For each image-question pair, we first convert the image $v_i$ into a textual context $c_i$ following \cite{yang2021empirical}, where $c_i$ consists of a caption generated from an image captioning model \cite{zhang2021vinvl} and a list of tags predicted by the public Microsoft Azure tagging API3\footnote{Azure Tagging API:\url{https://westus.dev.cognitive.microsoft.com/docs/services/computer-vision-v3-2/operations/56f91f2e778daf14a499f21b}}.
We then construct two carefully designed text prompts $Prompt_Q$ and $Prompt_{QC}$, where $Q$ stands for question and $QC$ stands for question and context. $Prompt_{QC}$ consists of a general instruction sentence: ``Please list all the possible answers to the question.'', the textual context, the question, and few-shot in-context examples.
The examples are context-question-answers triples taken from the training set that are most similar to the current image-question pair.
Since we want to generate all the possible answers, we use all the gold answers and connect them with ``or'' in the few-shot examples.
$Prompt_{Q}$ has similar components: a slightly different instruction sentence, the question, and few-shot examples of question-answers pairs. 

Following \cite{yang2021empirical,gui-etal-2022-kat}, we use 16-shot in-context examples and calculate the similarity scores using CLIP \cite{radford2021learning} embedding of the images and the questions.
We utilize the frozen PLM $g$ to generate outputs for both $Prompt_Q$ and $Prompt_{QC}$ as demonstrated in Figure \ref{fig:gen_prompt}.
The outputs are combined together to form the final answer choices $\hat{A_i} =\{\hat{a_{i0}}, \hat{a_{i1}}, \hat{a_{i2}}, ...\}$ for the current image-question pair. Our goal is to have $a_i \in \hat{A_i}$.

\subsection{Answer Selection}

Given $v_i$, $c_i$, $q_i$, $\hat{A_i}$, this step trains a model $p$ that selects $\hat{a_i}$ from $\hat{A_i}$. Our goal is for $p$ to output $a_i$ when $a_i \in \hat{A_i}$.

Before training $p$, we first generate chain-of-thought (CoT) \cite{wei2022chain} style rationales to help guide the selection process, with inspirations from \cite{AOKVQA}.
Specifically, a fixed prompt is pre-designed to generate CoT rationales, with details in Figure \ref{fig:cot} in Appendix \ref{sec:appendix}.

\begin{figure}[ht]
\centering
\vspace{0.5em}
\includegraphics[width=0.47\textwidth]{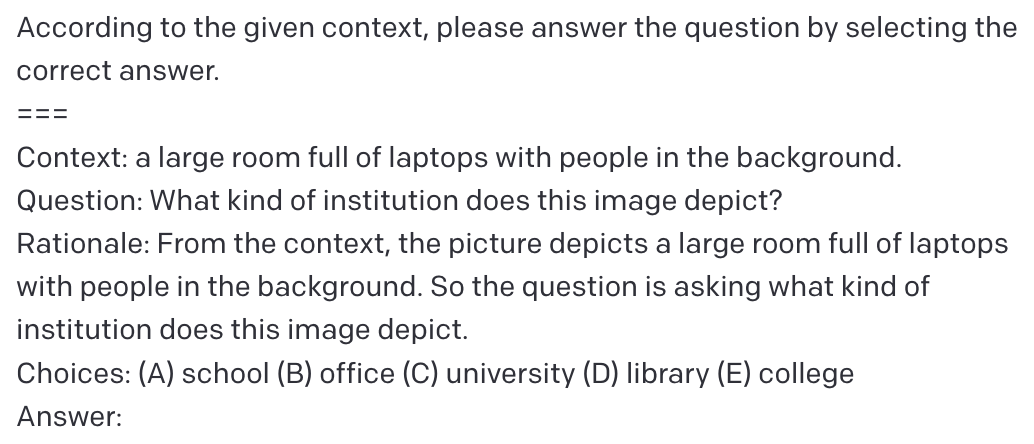}
\caption{Example input for the answer selection model for the image in \Cref{fig:example,fig:gen}.}
\vspace{-1.5em}
\label{fig:example:select}
\end{figure}

We then construct the input for the answer selection model.
In this paper, we plug in existing text generation models as $p$, and require them to output one choice with further fine-tuning on OK-VQA.
For each image-question pair, we concatenate the question $q_i$, the image -- represented by either $c_i$ or the image embedding using CLIP model \cite{radford2021learning}, the CoT rationale $cot_i$, and the generated answers choices $\hat{A_i}$. We also add sentinel tokens such that the input turns out to be in the following format: $Context: c_i$, $question: q_i$, $rationale: cot_i$, $choices: \hat{A_i}$, $answers:$ with minor adaptions for each specific $p$. Check Figure \ref{fig:example:select} for inference.

\section{Experiment}
\label{sec:exp}

% \begin{table*}[h]
% % \small
% \centering
% \setlength{\tabcolsep}{4pt}
% \begin{tabular}{c|cccc|c}
% \hline
% PLM& Top1 (\%)  & Top3 (\%)  & Top5 (\%)  & Top n (\%) & Avg $\#$ \\ \hline
% GPT J        &  37.23  & 53.04& 57.28& 58.26& 2.69    \\
% UL2   & 38.24  & 52.23& 54.18& 54.52& 3.25    \\
% GPT 3   & 48.0     &   -   &   -  &   -   &- \\
% OPT 175b   & 38.2   & 54.3 & 58.74& 64.04&??\\
% Codex       & 53.99  & 69.14& 73.73& 74.34& 4.54   \\
% \hline
% \end{tabular}
% \caption{\fxy{combine tables}}
% \label{tbl:exp:gen}
% \vspace{0.5em}
% \end{table*}

\begin{table*}[ht]
% \small
\centering
\setlength{\tabcolsep}{10pt}
\begin{tabular}{c|ccccc|c}
\hline
PLM& Prompt Type & Top1 (\%)  & Top3 (\%)  & Top5 (\%)  & All (\%) & Avg $\#$ \\ \hline
GPT J      & $Prompt_Q$  & 32.4 & 46.1 & 46.7 & 46.7 & 2.6    \\
     & $Prompt_{QC}$  & 37.1 & 49.5 & 50.7 & 50.7 & 3.0    \\
        & both & 37.1 & 52.0 & 55.9 & 57.1 & 4.1    \\
\hline
UL2        & $Prompt_Q$  & 32.6 & 45.4 & 46.4 & 46.5 & 2.7     \\
     & $Prompt_{QC}$  & 37.5 & 51.3 & 52.8 & 52.9 & 3.0    \\
 & both       & 37.5 & 53.1 & 57.0 & 58.0& 4.1    \\
\hline
GPT-3  & $Prompt_{QC}$  & 48.0     &   -   &  -   &   -   & -\\ 
\hline
OPT  & $Prompt_Q$  & 34.21    &   48.45   &   49.7  &   49.8   &3.0\\
  & $Prompt_{QC}$  &  37.8   &   52.9   &  55.0   &   55.4   &3.7\\
  & both    & 37.8   & 55.6 & 61.0 & 63.4 & 5.2\\
\hline
\textbf{Codex}   & $Prompt_Q$  & 44.8 & 58.8 & 59.8 & 59.8   & 3.1     \\
      & $Prompt_{QC}$ & 52.9 & 67.8 & 68.9 & 68.9   & 3.2 \\
  & both  & 52.9 & 68.6 & 72.6 & \textbf{73.5} & 4.5   \\
\hline
\textbf{ensembled}   & both  & 52.9 & 68.6 & 74.6 & \textbf{81.9} & 11.0   \\
\bottomrule
\end{tabular}
\caption{Answer choices generation result on OK-VQA, representing the external knowledge coverage.
Top 1, Top 3, Top 5, and All represent the highest accuracy that can be achieved using top 1, top 3, top 5, and all answer choices. All results are in accuracy scores evaluated following \cite{antol2015vqa}. ``both'' means that we combine the answer choices generated using both prompts. ``ensembled'' means that we combine the answer choices of all four PLMs. Note that the GPT-3 result is taken from \cite{yang2021empirical}.}
\label{tbl:exp:gen}
\vspace{-0.5em}
\end{table*}

\begin{table*}[ht]
\centering
\small
{
\vspace{-0.2cm}
\setlength{\tabcolsep}{1.3em}
    \begin{tabular}{c|l|c|c}
    \toprule[0.7pt]
    \textbf{Method} & External Knowledge Source & Answer Selector & Acc(\%) \\ 
    \hline
    MUTAN+AN \cite{ben2017mutan}  & Wiki & - & 27.8 \\
    ConceptBERT \cite{garderes-etal-2020-conceptbert} &ConceptNet&- & 33.7\\
    KRISP \cite{marino2021krisp}  & Wiki+ConceptNet & - &38.9 \\
    MAVEx \cite{wu2022multi}   &Wiki+ConceptNet+Google Images &- & 39.4 \\
    PiCA \cite{yang2021empirical}  & Frozen GPT-3 Wiki& - &48.0 \\
    % CBM \cite{salaberria2023image} & - & - & 47.9 \\
    KAT \cite{gui-etal-2022-kat} (ensemble)  &Wiki+Frozen GPT-3 Wiki& - &54.4\\
    \hline
    \hline
     ClipCap \cite{mokady2021clipcap} . & - & - &22.8\\
    \hline
    \hline
      & Frozen GPT-J &  &29.5\\
    & Frozen UL2 & & 33.1\\
      \modelname  & Frozen OPT & ClipCap  & 31.3\\
      & Frozen Codex &  &35.3\\
      & All 4 Frozen PLMs &  & 38.0\\
    \hline
    % \hline
     & Frozen GPT-J &  &29.6\\
    \modelname & Frozen UL2 & IterPLM &33.8\\
     & Frozen OPT &  &58.5\\
     & Frozen Codex &  &45.7\\
    % \hline
    \hline
     & Frozen GPT-J & &47.2\\
     & Frozen UL2 & &45.8\\
     \modelname& Frozen OPT & UnifiedQA &47.8\\
     & Frozen Codex & (ensemble) &51.2\\
     & All 4 Frozen PLMs & & 45.6\\
    % \hline
    \hline
     &Wiki+Frozen GPT-J &  &50.3\\
     &Wiki+Frozen UL2 & &52.2\\
     \textbf{\modelname} &Wiki+Frozen OPT & KAT  &53.0\\
     &Wiki+Frozen Codex & (ensemble) & \textbf{58.5}\\
     &Wiki+ All 4 Frozen PLMs & & 57.9\\
     % &Wiki+ Frozen Codex + GPT-3 & & 58.2\\
    \bottomrule[0.7pt]
    \end{tabular}
}
\caption{VQA results on the OK-VQA benchmark comparing to standard baselines. ``Wiki'' stands for ``Wikipedia'' and the ``Wiki'' resource in the last row's block is brought by the answer selector KAT. ``All 4 Frozen PLMs'' means that we use all the answer choices generated by GPT-J, UL2, OPT, and Codex. When we have UnifiedQA or KAT as answer selector, we train with 3 random seeds and denote the results as $ensemble$ following \cite{gui-etal-2022-kat}.}
% \vspace{-0.5cm}
\label{tbl:exp:select}
\end{table*}

\begin{table}[ht]
% \small
\setlength{\tabcolsep}{2pt}
\begin{tabular}{l||ccc}
\hline
\hline
KAT & Top1 & All w/o cot & All w/ cot\\ \hline \hline
GPT-J (single)   &  45.9  & 47.8 & 49.6 \\
GPT-J (ensemble) &  46.6 &  48.4 & 50.3\\
\hline
UL2 (single)    & 50.2 & 50.7  & 51.2 \\
UL2 (ensemble)  & 51.1 &  51.5 &  52.2\\
\hline
OPT (single)    & 51.7 & 52.3  & 52.5  \\
OPT (ensemble)  & 52.1 &  52.9 & 53.0\\
\hline
Codex (single)  &  56.2 & 57.1  & 57.5 \\
Codex (ensemble) &  57.1 & 58.1 & \textbf{58.5}\\ 
\hline
All (single)   &56.4 & 56.9 & 57.0 \\
All (ensemble) & 57.0 & 57.6 & 57.9 \\ 
\hline
% \hline
\end{tabular}
% \caption{Ablation study on how different inputs influence the answer selection result using KAT \cite{gui-etal-2022-kat} on OK-VQA. All results are in accuracy scores. ``Top1'' stands for using Top 1 answer choice, ``All'' means using all answer choices, and ``All w/ CoT'' means using all answer choices along with the CoT rationales. Following the original KAT method, we train our model with 3 random seeds and denote the average scores as $single$ and majority vote results as $ensemble$. The last two rows use combined answer choices from all four frozen PLMs.}
% \label{tbl:exp:kat}
\vspace{2em}
% \end{table}

% \begin{table}[h]
% \small
\setlength{\tabcolsep}{7pt}
\begin{tabular}{l||cc}
\hline
UnifiedQA  & All w/o cot  & All w/ cot \\ \hline\hline
GPT-J (single)   & 45.6 & 46.0       \\
GPT-J (ensemble) & 46.6 & 47.2       \\ \hline
UL2 (single)     & 44.8 & 44.6       \\
UL2 (ensemble)   & 45.8 & 45.8       \\ \hline
OPT (single)     & 47.9 & 46.8       \\
OPT (ensemble)   & 49.0 & 47.8       \\ \hline
Codex (single)   & 51.1 & 50.4       \\
Codex (ensemble) & \textbf{52.1} & 51.2       \\ \hline
All (single)     & 45.1 & 44.6       \\
All (ensemble)   & 45.7 & 45.3       \\ \hline
\end{tabular}
\caption{Ablation study investing how different inputs influence the answer selection results using KAT (top) and UnifiedQA (bottom) on OK-VQA in accuracy scores. ``Top1'' means using Top 1 answer choice,``All'' in the first row means using all answer choices, to form the input respectively. ``cot'' means the CoT rationales. We train with 3 random seeds and denote the average scores as $single$ and majority vote results as $ensemble$.``All'' in the leftmost column represent using combined answer choices from all four PLMs.}
\label{tbl:exp:unified}
\vspace{-0.5em}
\end{table}

% \begin{table*}[h]
% \begin{tabular}{l|lll}
% \hline
% UnifiedQA & No wiki & Top 5 Q + C candidates    & Top 5 Q + C candidate + rationale \\ \hline
% GPTJ      & 0.38    & 45.6 (0.2) --> 46.6 & 46.0 (0.2) --> 47.2\\
% UL2       & 0.39    & 44.8 (0.1) --> 45.8 & 44.6 (0.4) --> 45.8\\
% OPT       & 0.383   & 47.9 (0.5) -> 49.0   & 46.8 (0.4) --> 47.8\\ 
% Codex     & 0.55    & 51.1 (0.3) --> 52.1 & 50.4 (0.3) --> 51.2\\
% ALL     & -     & 45.1 (0.4) --> 45.7 & 44.6 (0.4) --> 45.3\\\hline
% \end{tabular}
% \caption{Main result of UnifiedQA, 3 repetitions and ensemble result}
% \label{tbl:exp:unified}
% \vspace{0.5em}
% \end{table*}

% \begin{table}[h]
% \setlength{\tabcolsep}{8pt}
% % \small
% \begin{tabular}{l|cccc}
% \hline
% & GPT-J & UL2 & OPT & Codex \\
% \hline
%  w/o cot  & 28.5 &   29.1   &  31.6   & 45.6    \\
% w/ cot & 28.1 & 32.3 &  33.5    & 44.9 \\
% \hline
% \end{tabular}
% \caption{Main result for answer selection using UnifiedQA, we use 3 random seeds and show the ensembled result.}
% \label{tbl:exp:iter}
% \vspace{0.5em}
% \end{table}

\begin{table}[ht]
\setlength{\tabcolsep}{8pt}
% \small
\begin{tabular}{l|cccc}
\hline
& GPT-J & UL2 & OPT & Codex \\
\hline
 w/o cot  & 28.5 &   29.1   &  31.6   & \textbf{45.6}    \\
w/ cot & 28.1 & 32.3 &  33.5    & 44.9 \\
\hline
\end{tabular}
\caption{Ablation study on how different inputs influence the answer selection result using IterPLM: iterative prompting using the same PLM, on OK-VQA. All results are in accuracy scores. 
% ``w/o cot'' means not using CoT rationales, and ``w/ CoT'' means using CoT rationales. 
Both setting use all the answer choices.}
\label{tbl:exp:iter}
\vspace{-0.5em}
\end{table}

\begin{table}[ht]
\setlength{\tabcolsep}{4pt}
% \small
\begin{tabular}{l|c|cccc}
\hline 
& Type & GPT-J & UL2 & OPT & Codex \\
\hline
 & DG  & \multicolumn{4}{c}{23.5} \\
ViT-L\_14 & w/o cot   & 28.7 &  30.3   &    29.1  & 33.4     \\
&  w/ cot & 29.5 & 33.1 &  31.3    & \textbf{35.3} \\ \hline
& DG   & \multicolumn{4}{c}{21.6}\\
RN50x64 & w/o cot  & 29.3 &  30.3   &    28.6    &  34.5      \\
&w/ cot & 29.6 & 32.6  &    31.4   &  \textbf{36.4} \\
\hline
\end{tabular}
\caption{Ablation study on how different inputs influence the answer selection result using ClipCapVQA \cite{mokady2021clipcap} on OK-VQA. The first column represents two CLIP checkpoints. ``DG'' represents direct generation without any answer choices.
% ``w/o cot'' means not using CoT rationales, and ``w/ CoT'' means using CoT rationales. All variants despite DG use all the answer choices.
}
\label{tbl:exp:clipcap}
\vspace{-0.5em}
\end{table}
\subsection{Dataset}
\textbf{OK-VQA} \cite{marino2021krisp} is a widely used VQA dataset that requires external world knowledge outside of the image to answer the question. The dataset contains 14,031 images from the COCO dataset \cite{lin2014microsoft} and 14,055 crowd-sourced questions covering a variety of knowledge categories, with 9,009 training data and 5,046 testing data. Each question has ten annotated answers (possibly repeated), and we follow the standard evaluation metric recommended by the VQA challenge \cite{antol2015vqa}.
The external knowledge required in OK-VQA is not provided and there is no designated external knowledge source (such as a knowledge base), leaving the benchmark more challenging.
% \textbf{A-OKVQA} \cite{AOKVQA} is a new knowledge-based visual question answering benchmark recently proposed. 
% A-OKVQA is an Augmented successor of OK-VQA and contains a diverse set of 25K questions requiring a broad base of commonsense and world knowledge to answer. Questions in A-OKVQA are challenging, conceptually diverse, require knowledge outside the image, and in contrast to existing knowledge-based visual question answering datasets, they cannot be answered by simply querying a knowledge base. 
\subsection{Publicly Available PLMs}
We experiment with four different-sized PLMs that are publicly available as follows:\\
\textbf{Codex} \cite{chen2021evaluating}
    The Codex models are descendants of GPT-3 models that can understand and generate code. Their training data contains both natural language and billions of lines of public code from GitHub. We use the version $code-davinci-002$ of Codex.\\
 \textbf{OPT-175b} \cite{zhang2022opt}
    Open Pre-trained Transformers (OPT) is a suite of decoder-only pre-trained transformers ranging from 125M to 175B parameters trained on publicly available datasets. We use the version 175 billion parameters of OPT.\\
 \textbf{UL2} \cite{tay2022unifying} 
    Unified Language Learner (UL2) is 20 billion parameter novel language pre-training paradigm that improves the performance of language models universally across datasets and setups released recently. UL2 frames different objective functions for training language models as denoising tasks, where the model has to recover missing sub-sequences of a given input. 
    % During pre-training it uses a novel mixture-of-denoisers that samples from a varied set of such objectives, each with different configurations.
    \\
 \textbf{GPT-J} \cite{gpt-j}
    GPT-J is a 6 billion parameter, autoregressive text generation model trained following \cite{mesh-transformer-jax}. The model consists of 28 layers with a model dimension of 4096, and a feed-forward dimension of 16384.\\ 
During prompting, we always set the temperature to 0.001 and max token to 15.

\subsection{Answer Choices Generation Results}
The answer choice generation result is shown in Table \ref{tbl:exp:gen}. Top 1, Top 3,..., All represent the highest accuracy that can be achieved using top 1, top 3, ..., and all answer choices, calculated following the standard VQA evaluation metric in \cite{antol2015vqa}.
Note that the GPT-3 score is taken from \cite{yang2021empirical}. We do not experiment with GPT-3 in this paper due to the required cost. Avg \# stands for the average number of answer choices.

While previous VQA methods also retrieve from PLMs, they have a similar result as if using $Prompt_{QC}$ and Top1 choice.
As discussed before, these generation results can represent the external knowledge coverage ratio.
From the table, Codex covers the majority of the knowledge needed and has the highest score of 73.5\%.
Using our prompt-engineering method, the knowledge coverages of all PLMs increase by a large margin of at least 20\% (which are the accuracy differences between Top1 choice by $Prompt_{QC}$ and All choices by both prompts).

\subsection{Answer Selection Models}
\label{sec:exp:select}
We plug in existing text-generation models as answer selectors and experiment on four methods:\\
\textbf{KAT} \cite{gui-etal-2022-kat} is a VQA method that uses a sequence-to-sequence model composed of an encoder and a decoder, similar to T5 \cite{raffel2020exploring}. As far as this paper is written, KAT is known to be the SOTA method on OK-VQA benchmark. \\
\textbf{ClipCap} \cite{mokady2021clipcap} uses the CLIP \cite{radford2021learning} encoding as a prefix to generate textual captions by employing a simple mapping network over the raw encoding, and then fine-tunes a language model to generate a valid caption. The language model we use here is GPT-2. In this paper, we adapt this model by adding question tokens, CoT rationale tokens, and answer choices tokens to the prefix as input, with the target to generate answers instead of captions. We train the mapping network from scratch and also fine-tune GPT-2.\\
\textbf{IterPLM} Inspired by previous work \cite{wang2022shepherd}, we use iterative prompting with the same PLM in choice generation for correct answer selection. A snippet of an example prompt is shown in Figure \ref{fig:example:select}. We use 8-shot in-domain examples with the temperature set to 0.001 and max token set to 5.\\
\textbf{UnifiedQA} \cite{khashabi2022unifiedqa,khashabi-etal-2020-unifiedqa} is a multiple-choice question answering (QA) model that performs well across 20 QA datasets, using the T5ForConditionalGeneration model. We load UnifiedQA v2 \cite{khashabi2022unifiedqa} checkpoint unifiedqa-v2-t5-large-1251000.

\subsection{End-task VQA Results}
As illustrated in Table \ref{tbl:exp:select}, we compare our proposed pipeline against several standard baseline approaches: MUTAN+AN \cite{ben2017mutan}, ConceptBERT \cite{garderes-etal-2020-conceptbert}, KRISP \cite{marino2021krisp}, MAVEx \cite{wu2022multi}, PiCA \cite{yang2021empirical}, and KAT \cite{gui-etal-2022-kat}, on the OK-VQA data test set.
\modelname outperforms the previous SOTA by an absolute 4\% margin, achieving the new SOTA. 

Comparing different answer selectors, it is surprising that the two transformer-based text-only models: UnifiedQA and KAT significantly outperform the multi-modal ClipCap model by around 20\% on average, even though their sizes (T5 large) are much smaller than that of GPT-2.
We believe this phenomenon is because the Clip image embeddings trained using image captions do not have enough granularity to support reasoning over the image, question, and answer choices for answer selection, compared to T5 models.
Besides, IterPLM has much worse scores than we imagined. While many papers \cite{wang2022shepherd} show that iterative prompting should boost the performance, our experiments suggest that asking the PLMs to select between their own output at the highest confidence is indeed a very difficult problem for them. 

In Table \ref{tbl:exp:select}, we also compare single PLM answer choices with ensembled choices by all four PLMs, with the latter showing lower scores. We believe this is because the answer selectors we experiment on are not good enough, and thus increasing choice numbers turns out to
% can in contrast bring in noise and therefore
hurt the performance.

\subsection{Implementation Details}
In the answer choice generation step, we use 16-shot in-context examples on the test data. On the training data, because we have ten gold answers with repetitions, we use 4-shot in-context learning for faster generation. The temperature for PLM generation is set to be 0.001. The generation max token length is set to be 15. 
All experiments of selection models have been run in 8 NVIDIA V100 Tensor Core GPUs with 32 GiB of memory each, 96 custom Intel Xeon Scalable (Skylake) vCPUs, and 1.8 TB of local NVMe-based SSD storage.
The running times for KAT, UnifiedQA and ClipCap are less than 4, 2 and 1 hours, respectively.
OPT-175b model is locally set up in 32 NVIDIA V100 Tensor Core GPUs to make inferences.
The learning rates for KAT, UnifiedQA and Clipcap are set as 3e-5, 5e-5 and 2e-5, respectively, for all experiments. Optimizer AdamW \citep{loshchilov2017decoupled} is used for all selection models.

\section{Ablation Studies}
% \label{sec:abla}
We perform qualitative and quantitative analysis
on the answer selection results to better understand whether the expanded external knowledge coverage benefits the end-task VQA much.
As illustrated in \Cref{tbl:exp:unified,tbl:exp:iter,tbl:exp:clipcap},
we investigate the impact of various inputs on the answer selection results, with answer choices representing the retrieved  knowledge.

\textbf{CoT Rationale Impact} From the experiments results in \Cref{tbl:exp:unified,tbl:exp:iter,tbl:exp:clipcap} where we compare the settings: ``w/cot'' and ``w/o cot'', input with CoT rationales consistently boosts the answer selection performance of KAT, UnifiedQA, and ClipCap. However, this conclusion fails for iterative prompting -- adding CoT hurts the performance of IterPLM when we use GPT-J and Codex. We believe this can result from the difference in CoT qualities, and different pre-training methods and data.

\textbf{Choice Number Impact}
As shown in Table \ref{tbl:exp:unified}, larger knowledge coverage, represented by using choices from all four PLMs versus a single PLM, can not consistently increase the performance of KAT or UnifiedQA. As we compare the results on Codex choices and that on all PLMs choices, more choices always lead to lower accuracy scores. This is somehow against our instinct, and we believe it is because our answer selectors are not good enough. 
Digging deeper into the problem, we further compare the difference between using Top1 choices and all choices in KAT as in the top table. Note that the Top1 results here are not the same as the Top1 accuracy in Table \ref{tbl:exp:gen} because KAT uses Wikipedia knowledge by design so it further expands knowledge coverage.
We can see that using all choices is consistently better than using Top 1 choice. However, the improvements are too small (0.4-1.9 \%) considering that their knowledge coverages differ by at least 20\% as in Table \ref{tbl:gen}, suggesting that KAT, while being the best, is still not the ideal selection model, and motivating future research in this direction. 

\textbf{Multi-modal Selector Impact}
As demonstrated in Table \ref{tbl:exp:clipcap}, we experiment with the two versions of CLIP embedding: ``ViT-L\_14'' and ``RN50x64'' and the difference between direct generation (DG) and answer selection is constantly large -- providing answer choices definitely helps ClipCap to generate the correct answer.

\textbf{Ensemble Impact}
Our answer choice generation step is indeed ensembling on PLMs results. Previous VQA methods that retrieve from PLMs also conduct ensembling but in a different way \cite{yang2021empirical}. They usually request the same prompt (see example in Figure \ref{fig:gen_prompt}) multiple times and take the majority-voted answer. This process is called multi-query ensemble, and could boost the GPT-3 performance by about 5\%. We argue that our proposed \modelname prompting is superior to multi-query ensemble in that we allow more diversity in the output and provide VQA systems more explainability by separating the choice-generation and selection steps, without additional API request cost or longer inference time. 

% \textbf{Explainability}

% \textbf{Comparison with traditional VQA methods}

% \fxy{add 3 examples: no cover, cover not select correctly, cover and select correctly. Add error analysis.}

% \input{tables/implementation_detail.tex}
\section{Conclusion}
In this paper, we propose \modelname: a new VQA pipeline following a generate-then-select strategy guided by world knowledge. \modelname proposes a new prompting method that largely increases the external knowledge coverage by a margin of more than 20\% compared to previous approaches on the OK-VQA benchmark. Our pipeline achieves the new SOTA 58.5\% on the end-task performance
% The ablation studies suggest that the answers selection models we experiment with are all not ideal, and cannot benefit from the expanded world knowledge
, encouraging avenues for future studies.
\section{Limitations}
While the previous VQA methods that retrieve from PLMs all use GPT-3, we do not experiment with GPT-3 in this paper due to the additional cost.
We only focus on applying text-generation models as answer selectors, while classification models could also potentially be good answer selectors. 
The multi-modal CLIP embedding has already been surpassed by several recent studies \cite{alayrac2022flamingo,singh2022flava,lu2022unified} and therefore ClipCap cannot represent the performance of multi-modal answer selectors.

\section{Ethical Considerations }
The authors of this paper acknowledge the significance of responsible NLP in research and development. The objective of this research is to enhance the capabilities of visual question answering models, guided by human values-based world knowledge. We strive to ensure that the model is not only accurate and efficient, but also fair and unbiased. We recognize that the VQA technology may have a substantial impact on society and pledge to be transparent in sharing our findings and progress with relevant users and stakeholders.

\section*{Acknowledgments}
The authors would like to thank researchers at AWS AI Labs who commented on or otherwise supported throughout the course of this project, including Simeng Han,  Donghan Yu, Sijia Wang, and Shuaichen Chang.

\newpage

\newpage

\bibliography{anthology,custom}

\begin{thebibliography}{51}
\expandafter\ifx\csname natexlab\endcsname\relax\def\natexlab#1{#1}\fi

\bibitem[{Alayrac et~al.(2022)Alayrac, Donahue, Luc, Miech, Barr, Hasson, Lenc,
  Mensch, Millican, Reynolds et~al.}]{alayrac2022flamingo}
Jean-Baptiste Alayrac, Jeff Donahue, Pauline Luc, Antoine Miech, Iain Barr,
  Yana Hasson, Karel Lenc, Arthur Mensch, Katie Millican, Malcolm Reynolds,
  et~al. 2022.
\newblock Flamingo: a visual language model for few-shot learning.
\newblock \emph{arXiv preprint arXiv:2204.14198}.

\bibitem[{Anderson et~al.(2018)Anderson, He, Buehler, Teney, Johnson, Gould,
  and Zhang}]{anderson2018bottom}
Peter Anderson, Xiaodong He, Chris Buehler, Damien Teney, Mark Johnson, Stephen
  Gould, and Lei Zhang. 2018.
\newblock Bottom-up and top-down attention for image captioning and visual
  question answering.
\newblock In \emph{Proceedings of the IEEE conference on computer vision and
  pattern recognition}, pages 6077--6086.

\bibitem[{Antol et~al.(2015)Antol, Agrawal, Lu, Mitchell, Batra, Zitnick, and
  Parikh}]{antol2015vqa}
Stanislaw Antol, Aishwarya Agrawal, Jiasen Lu, Margaret Mitchell, Dhruv Batra,
  C~Lawrence Zitnick, and Devi Parikh. 2015.
\newblock Vqa: Visual question answering.
\newblock In \emph{Proceedings of the IEEE international conference on computer
  vision}, pages 2425--2433.

\bibitem[{Ben-Younes et~al.(2017)Ben-Younes, Cadene, Cord, and
  Thome}]{ben2017mutan}
Hedi Ben-Younes, R{\'e}mi Cadene, Matthieu Cord, and Nicolas Thome. 2017.
\newblock Mutan: Multimodal tucker fusion for visual question answering.
\newblock In \emph{Proceedings of the IEEE international conference on computer
  vision}, pages 2612--2620.

\bibitem[{Biten et~al.(2022)Biten, Litman, Xie, Appalaraju, and
  Manmatha}]{Biten_2022_CVPR}
Ali~Furkan Biten, Ron Litman, Yusheng Xie, Srikar Appalaraju, and R.~Manmatha.
  2022.
\newblock Latr: Layout-aware transformer for scene-text vqa.
\newblock In \emph{Proceedings of the IEEE/CVF Conference on Computer Vision
  and Pattern Recognition (CVPR)}, pages 16548--16558.

\bibitem[{Brown et~al.(2020)Brown, Mann, Ryder, Subbiah, Kaplan, Dhariwal,
  Neelakantan, Shyam, Sastry, Askell et~al.}]{brown2020language}
Tom Brown, Benjamin Mann, Nick Ryder, Melanie Subbiah, Jared~D Kaplan, Prafulla
  Dhariwal, Arvind Neelakantan, Pranav Shyam, Girish Sastry, Amanda Askell,
  et~al. 2020.
\newblock Language models are few-shot learners.
\newblock \emph{Advances in neural information processing systems},
  33:1877--1901.

\bibitem[{Chen et~al.(2021)Chen, Tworek, Jun, Yuan, Pinto, Kaplan, Edwards,
  Burda, Joseph, Brockman et~al.}]{chen2021evaluating}
Mark Chen, Jerry Tworek, Heewoo Jun, Qiming Yuan, Henrique Ponde de~Oliveira
  Pinto, Jared Kaplan, Harri Edwards, Yuri Burda, Nicholas Joseph, Greg
  Brockman, et~al. 2021.
\newblock Evaluating large language models trained on code.
\newblock \emph{arXiv preprint arXiv:2107.03374}.

\bibitem[{Chen et~al.(2022)Chen, Wang, Changpinyo, Piergiovanni, Padlewski,
  Salz, Goodman, Grycner, Mustafa, Beyer et~al.}]{chen2022pali}
Xi~Chen, Xiao Wang, Soravit Changpinyo, AJ~Piergiovanni, Piotr Padlewski,
  Daniel Salz, Sebastian Goodman, Adam Grycner, Basil Mustafa, Lucas Beyer,
  et~al. 2022.
\newblock Pali: A jointly-scaled multilingual language-image model.
\newblock \emph{arXiv preprint arXiv:2209.06794}.

\bibitem[{Chowdhery et~al.(2022)Chowdhery, Narang, Devlin, Bosma, Mishra,
  Roberts, Barham, Chung, Sutton, Gehrmann et~al.}]{chowdhery2022palm}
Aakanksha Chowdhery, Sharan Narang, Jacob Devlin, Maarten Bosma, Gaurav Mishra,
  Adam Roberts, Paul Barham, Hyung~Won Chung, Charles Sutton, Sebastian
  Gehrmann, et~al. 2022.
\newblock Palm: Scaling language modeling with pathways.
\newblock \emph{arXiv preprint arXiv:2204.02311}.

\bibitem[{Fu et~al.(2022)Fu, Zhou, Chandratreya, Vondrick, and
  Roth}]{fu-etal-2022-theres}
Xingyu Fu, Ben Zhou, Ishaan Chandratreya, Carl Vondrick, and Dan Roth. 2022.
\newblock \href {https://doi.org/10.18653/v1/2022.acl-long.81} {There{'}s a
  time and place for reasoning beyond the image}.
\newblock In \emph{Proceedings of the 60th Annual Meeting of the Association
  for Computational Linguistics (Volume 1: Long Papers)}, pages 1138--1149,
  Dublin, Ireland. Association for Computational Linguistics.

\bibitem[{Gao et~al.(2022)Gao, Ping, Thattai, Reganti, Wu, and
  Natarajan}]{Gap2022}
Feng Gao, Qing Ping, Govind Thattai, Aishwarya Reganti, Ying~Nian Wu, and Prem
  Natarajan. 2022.
\newblock \href
  {https://www.amazon.science/publications/transform-retrieve-generate-natural-language-centric-outside-knowledge-visual-question-answering}
  {Transform-retrieve-generate: Natural language-centric outside-knowledge
  visual question answering}.
\newblock In \emph{CVPR 2022}.

\bibitem[{Gard{\`e}res et~al.(2020)Gard{\`e}res, Ziaeefard, Abeloos, and
  Lecue}]{garderes-etal-2020-conceptbert}
Fran{\c{c}}ois Gard{\`e}res, Maryam Ziaeefard, Baptiste Abeloos, and Freddy
  Lecue. 2020.
\newblock \href {https://doi.org/10.18653/v1/2020.findings-emnlp.44}
  {{C}oncept{B}ert: Concept-aware representation for visual question
  answering}.
\newblock In \emph{Findings of the Association for Computational Linguistics:
  EMNLP 2020}, pages 489--498, Online. Association for Computational
  Linguistics.

\bibitem[{Goyal et~al.(2017)Goyal, Khot, Summers{-}Stay, Batra, and
  Parikh}]{balanced_vqa_v2}
Yash Goyal, Tejas Khot, Douglas Summers{-}Stay, Dhruv Batra, and Devi Parikh.
  2017.
\newblock Making the {V} in {VQA} matter: Elevating the role of image
  understanding in {V}isual {Q}uestion {A}nswering.
\newblock In \emph{Conference on Computer Vision and Pattern Recognition
  (CVPR)}.

\bibitem[{Gui et~al.(2022)Gui, Wang, Huang, Hauptmann, Bisk, and
  Gao}]{gui-etal-2022-kat}
Liangke Gui, Borui Wang, Qiuyuan Huang, Alexander Hauptmann, Yonatan Bisk, and
  Jianfeng Gao. 2022.
\newblock \href {https://doi.org/10.18653/v1/2022.naacl-main.70} {{KAT}: A
  knowledge augmented transformer for vision-and-language}.
\newblock In \emph{Proceedings of the 2022 Conference of the North American
  Chapter of the Association for Computational Linguistics: Human Language
  Technologies}, pages 956--968, Seattle, United States. Association for
  Computational Linguistics.

\bibitem[{Hu et~al.(2022)Hu, Hua, Yang, Shi, Smith, and Luo}]{hu2022promptcap}
Yushi* Hu, Hang* Hua, Zhengyuan Yang, Weijia Shi, Noah~A Smith, and Jiebo Luo.
  2022.
\newblock Promptcap: Prompt-guided task-aware image captioning.
\newblock \emph{arXiv preprint arXiv:2211.09699}.

\bibitem[{Jiang et~al.(2020)Jiang, Misra, Rohrbach, Learned-Miller, and
  Chen}]{jiang2020defense}
Huaizu Jiang, Ishan Misra, Marcus Rohrbach, Erik Learned-Miller, and Xinlei
  Chen. 2020.
\newblock In defense of grid features for visual question answering.
\newblock In \emph{Proceedings of the IEEE/CVF Conference on Computer Vision
  and Pattern Recognition}, pages 10267--10276.

\bibitem[{Khashabi et~al.(2022)Khashabi, Kordi, and
  Hajishirzi}]{khashabi2022unifiedqa}
Daniel Khashabi, Yeganeh Kordi, and Hannaneh Hajishirzi. 2022.
\newblock Unifiedqa-v2: Stronger generalization via broader cross-format
  training.
\newblock \emph{arXiv preprint arXiv:2202.12359}.

\bibitem[{Khashabi et~al.(2020)Khashabi, Min, Khot, Sabharwal, Tafjord, Clark,
  and Hajishirzi}]{khashabi-etal-2020-unifiedqa}
Daniel Khashabi, Sewon Min, Tushar Khot, Ashish Sabharwal, Oyvind Tafjord,
  Peter Clark, and Hannaneh Hajishirzi. 2020.
\newblock \href {https://doi.org/10.18653/v1/2020.findings-emnlp.171}
  {{UNIFIEDQA}: Crossing format boundaries with a single {QA} system}.
\newblock In \emph{Findings of the Association for Computational Linguistics:
  EMNLP 2020}, pages 1896--1907, Online. Association for Computational
  Linguistics.

\bibitem[{Li et~al.(2023)Li, Li, Savarese, and Hoi}]{li2023blip2}
Junnan Li, Dongxu Li, Silvio Savarese, and Steven Hoi. 2023.
\newblock {BLIP-2:} bootstrapping language-image pre-training with frozen image
  encoders and large language models.
\newblock In \emph{ICML}.

\bibitem[{Lin et~al.(2022)Lin, Xie, Chen, Xu, Zhu, and Yuan}]{lin2022revive}
Leroy Lin, Yujia Xie, Dongdong Chen, Yichong Xu, Chenguang Zhu, and Lu~Yuan.
  2022.
\newblock \href {https://openreview.net/forum?id=wwyiEyK-G5D} {{REVIVE}:
  Regional visual representation matters in knowledge-based visual question
  answering}.
\newblock In \emph{Advances in Neural Information Processing Systems}.

\bibitem[{Lin et~al.(2014)Lin, Maire, Belongie, Hays, Perona, Ramanan,
  Doll{\'a}r, and Zitnick}]{lin2014microsoft}
Tsung-Yi Lin, Michael Maire, Serge Belongie, James Hays, Pietro Perona, Deva
  Ramanan, Piotr Doll{\'a}r, and C~Lawrence Zitnick. 2014.
\newblock Microsoft coco: Common objects in context.
\newblock In \emph{European conference on computer vision}, pages 740--755.
  Springer.

\bibitem[{Lin and Byrne(2022)}]{lin-byrne-2022-retrieval}
Weizhe Lin and Bill Byrne. 2022.
\newblock \href {https://aclanthology.org/2022.emnlp-main.772} {Retrieval
  augmented visual question answering with outside knowledge}.
\newblock In \emph{Proceedings of the 2022 Conference on Empirical Methods in
  Natural Language Processing}, pages 11238--11254, Abu Dhabi, United Arab
  Emirates. Association for Computational Linguistics.

\bibitem[{Liu and Singh(2004)}]{liu2004conceptnet}
Hugo Liu and Push Singh. 2004.
\newblock Conceptnet—a practical commonsense reasoning tool-kit.
\newblock \emph{BT technology journal}, 22(4):211--226.

\bibitem[{Loshchilov and Hutter(2017)}]{loshchilov2017decoupled}
Ilya Loshchilov and Frank Hutter. 2017.
\newblock Decoupled weight decay regularization.
\newblock \emph{arXiv preprint arXiv:1711.05101}.

\bibitem[{Lu et~al.(2019)Lu, Batra, Parikh, and Lee}]{lu2019vilbert}
Jiasen Lu, Dhruv Batra, Devi Parikh, and Stefan Lee. 2019.
\newblock Vilbert: Pretraining task-agnostic visiolinguistic representations
  for vision-and-language tasks.
\newblock \emph{Advances in neural information processing systems}, 32.

\bibitem[{Lu et~al.(2022)Lu, Clark, Zellers, Mottaghi, and
  Kembhavi}]{lu2022unified}
Jiasen Lu, Christopher Clark, Rowan Zellers, Roozbeh Mottaghi, and Aniruddha
  Kembhavi. 2022.
\newblock Unified-io: A unified model for vision, language, and multi-modal
  tasks.
\newblock \emph{arXiv preprint arXiv:2206.08916}.

\bibitem[{Lu et~al.(2016)Lu, Yang, Batra, and Parikh}]{lu2016hierarchical}
Jiasen Lu, Jianwei Yang, Dhruv Batra, and Devi Parikh. 2016.
\newblock Hierarchical question-image co-attention for visual question
  answering.
\newblock \emph{Advances in neural information processing systems}, 29.

\bibitem[{Marino et~al.(2021)Marino, Chen, Parikh, Gupta, and
  Rohrbach}]{marino2021krisp}
Kenneth Marino, Xinlei Chen, Devi Parikh, Abhinav Gupta, and Marcus Rohrbach.
  2021.
\newblock Krisp: Integrating implicit and symbolic knowledge for open-domain
  knowledge-based vqa.
\newblock In \emph{Proceedings of the IEEE/CVF Conference on Computer Vision
  and Pattern Recognition}, pages 14111--14121.

\bibitem[{Marino et~al.(2019)Marino, Rastegari, Farhadi, and
  Mottaghi}]{marino2019ok}
Kenneth Marino, Mohammad Rastegari, Ali Farhadi, and Roozbeh Mottaghi. 2019.
\newblock Ok-vqa: A visual question answering benchmark requiring external
  knowledge.
\newblock In \emph{Proceedings of the IEEE/cvf conference on computer vision
  and pattern recognition}, pages 3195--3204.

\bibitem[{Mokady et~al.(2021)Mokady, Hertz, and Bermano}]{mokady2021clipcap}
Ron Mokady, Amir Hertz, and Amit~H Bermano. 2021.
\newblock Clipcap: Clip prefix for image captioning.
\newblock \emph{arXiv preprint arXiv:2111.09734}.

\bibitem[{Park et~al.(2020)Park, Bhagavatula, Mottaghi, Farhadi, and
  Choi}]{park2020visualcomet}
Jae~Sung Park, Chandra Bhagavatula, Roozbeh Mottaghi, Ali Farhadi, and Yejin
  Choi. 2020.
\newblock Visualcomet: Reasoning about the dynamic context of a still image.
\newblock In \emph{European Conference on Computer Vision}, pages 508--524.
  Springer.

\bibitem[{Radford et~al.(2021)Radford, Kim, Hallacy, Ramesh, Goh, Agarwal,
  Sastry, Askell, Mishkin, Clark et~al.}]{radford2021learning}
Alec Radford, Jong~Wook Kim, Chris Hallacy, Aditya Ramesh, Gabriel Goh,
  Sandhini Agarwal, Girish Sastry, Amanda Askell, Pamela Mishkin, Jack Clark,
  et~al. 2021.
\newblock Learning transferable visual models from natural language
  supervision.
\newblock In \emph{International Conference on Machine Learning}, pages
  8748--8763. PMLR.

\bibitem[{Raffel et~al.(2020)Raffel, Shazeer, Roberts, Lee, Narang, Matena,
  Zhou, Li, Liu et~al.}]{raffel2020exploring}
Colin Raffel, Noam Shazeer, Adam Roberts, Katherine Lee, Sharan Narang, Michael
  Matena, Yanqi Zhou, Wei Li, Peter~J Liu, et~al. 2020.
\newblock Exploring the limits of transfer learning with a unified text-to-text
  transformer.
\newblock \emph{J. Mach. Learn. Res.}, 21(140):1--67.

\bibitem[{Salaberria et~al.(2023)Salaberria, Azkune, de~Lacalle, Soroa, and
  Agirre}]{salaberria2023image}
Ander Salaberria, Gorka Azkune, Oier~Lopez de~Lacalle, Aitor Soroa, and Eneko
  Agirre. 2023.
\newblock Image captioning for effective use of language models in
  knowledge-based visual question answering.
\newblock \emph{Expert Systems with Applications}, 212:118669.

\bibitem[{Schwenk et~al.(2022)Schwenk, Khandelwal, Clark, Marino, and
  Mottaghi}]{AOKVQA}
Dustin Schwenk, Apoorv Khandelwal, Christopher Clark, Kenneth Marino, and
  Roozbeh Mottaghi. 2022.
\newblock A-okvqa: A benchmark for visual question answering using world
  knowledge.
\newblock \emph{arXiv}.

\bibitem[{Shih et~al.(2016)Shih, Singh, and Hoiem}]{shih2016look}
Kevin~J Shih, Saurabh Singh, and Derek Hoiem. 2016.
\newblock Where to look: Focus regions for visual question answering.
\newblock In \emph{Proceedings of the IEEE conference on computer vision and
  pattern recognition}, pages 4613--4621.

\bibitem[{Singh et~al.(2022)Singh, Hu, Goswami, Couairon, Galuba, Rohrbach, and
  Kiela}]{singh2022flava}
Amanpreet Singh, Ronghang Hu, Vedanuj Goswami, Guillaume Couairon, Wojciech
  Galuba, Marcus Rohrbach, and Douwe Kiela. 2022.
\newblock Flava: A foundational language and vision alignment model.
\newblock In \emph{Proceedings of the IEEE/CVF Conference on Computer Vision
  and Pattern Recognition}, pages 15638--15650.

\bibitem[{Singh et~al.(2019)Singh, Natarajan, Shah, Jiang, Chen, Batra, Parikh,
  and Rohrbach}]{singh2019towards}
Amanpreet Singh, Vivek Natarajan, Meet Shah, Yu~Jiang, Xinlei Chen, Dhruv
  Batra, Devi Parikh, and Marcus Rohrbach. 2019.
\newblock Towards vqa models that can read.
\newblock In \emph{Proceedings of the IEEE/CVF conference on computer vision
  and pattern recognition}, pages 8317--8326.

\bibitem[{Tay et~al.(2022)Tay, Dehghani, Tran, Garcia, Bahri, Schuster, Zheng,
  Houlsby, and Metzler}]{tay2022unifying}
Yi~Tay, Mostafa Dehghani, Vinh~Q Tran, Xavier Garcia, Dara Bahri, Tal Schuster,
  Huaixiu~Steven Zheng, Neil Houlsby, and Donald Metzler. 2022.
\newblock Unifying language learning paradigms.
\newblock \emph{arXiv preprint arXiv:2205.05131}.

\bibitem[{Wang(2021)}]{mesh-transformer-jax}
Ben Wang. 2021.
\newblock {Mesh-Transformer-JAX: Model-Parallel Implementation of Transformer
  Language Model with JAX}.
\newblock \url{https://github.com/kingoflolz/mesh-transformer-jax}.

\bibitem[{Wang and Komatsuzaki(2021)}]{gpt-j}
Ben Wang and Aran Komatsuzaki. 2021.
\newblock {GPT-J-6B: A 6 Billion Parameter Autoregressive Language Model}.
\newblock \url{https://github.com/kingoflolz/mesh-transformer-jax}.

\bibitem[{Wang et~al.(2022)Wang, Deng, and Sun}]{wang2022shepherd}
Boshi Wang, Xiang Deng, and Huan Sun. 2022.
\newblock Shepherd pre-trained language models to develop a train of thought:
  An iterative prompting approach.
\newblock \emph{In Proceedings of the 2022 Conference on Empirical Methods in
  Natural Language Processing}.

\bibitem[{Wang et~al.(2017)Wang, Wu, Shen, Dick, and Van
  Den~Hengel}]{wang2017fvqa}
Peng Wang, Qi~Wu, Chunhua Shen, Anthony Dick, and Anton Van Den~Hengel. 2017.
\newblock Fvqa: Fact-based visual question answering.
\newblock \emph{IEEE transactions on pattern analysis and machine
  intelligence}, 40(10):2413--2427.

\bibitem[{Wei et~al.(2022)Wei, Wang, Schuurmans, Bosma, brian ichter, Xia, Chi,
  Le, and Zhou}]{wei2022chain}
Jason Wei, Xuezhi Wang, Dale Schuurmans, Maarten Bosma, brian ichter, Fei Xia,
  Ed~H. Chi, Quoc~V Le, and Denny Zhou. 2022.
\newblock \href {https://openreview.net/forum?id=_VjQlMeSB_J} {Chain of thought
  prompting elicits reasoning in large language models}.
\newblock In \emph{Advances in Neural Information Processing Systems}.

\bibitem[{{Wikipedia contributors}(2004)}]{wiki}
{Wikipedia contributors}. 2004.
\newblock \href
  {https://en.wikipedia.org/w/index.php?title=Plagiarism&oldid=5139350}
  {Plagiarism --- {W}ikipedia{,} the free encyclopedia}.
\newblock [Online; accessed 22-July-2004].

\bibitem[{Wu et~al.(2022)Wu, Lu, Sabharwal, and Mottaghi}]{wu2022multi}
Jialin Wu, Jiasen Lu, Ashish Sabharwal, and Roozbeh Mottaghi. 2022.
\newblock Multi-modal answer validation for knowledge-based vqa.
\newblock In \emph{Proceedings of the AAAI Conference on Artificial
  Intelligence}, pages 2712--2721.

\bibitem[{Yang et~al.(2022)Yang, Gan, Wang, Hu, Lu, Liu, and
  Wang}]{yang2021empirical}
Zhengyuan Yang, Zhe Gan, Jianfeng Wang, Xiaowei Hu, Yumao Lu, Zicheng Liu, and
  Lijuan Wang. 2022.
\newblock An empirical study of gpt-3 for few-shot knowledge-based vqa.
\newblock In \emph{AAAI}.

\bibitem[{Yu et~al.(2019)Yu, Yu, Cui, Tao, and Tian}]{yu2019deep}
Zhou Yu, Jun Yu, Yuhao Cui, Dacheng Tao, and Qi~Tian. 2019.
\newblock Deep modular co-attention networks for visual question answering.
\newblock In \emph{Proceedings of the IEEE/CVF conference on computer vision
  and pattern recognition}, pages 6281--6290.

\bibitem[{Zellers et~al.(2019)Zellers, Bisk, Farhadi, and
  Choi}]{zellers2019vcr}
Rowan Zellers, Yonatan Bisk, Ali Farhadi, and Yejin Choi. 2019.
\newblock From recognition to cognition: Visual commonsense reasoning.
\newblock In \emph{The IEEE Conference on Computer Vision and Pattern
  Recognition (CVPR)}.

\bibitem[{Zhang et~al.(2021)Zhang, Li, Hu, Yang, Zhang, Wang, Choi, and
  Gao}]{zhang2021vinvl}
Pengchuan Zhang, Xiujun Li, Xiaowei Hu, Jianwei Yang, Lei Zhang, Lijuan Wang,
  Yejin Choi, and Jianfeng Gao. 2021.
\newblock Vinvl: Making visual representations matter in vision-language
  models.
\newblock \emph{CVPR 2021}.

\bibitem[{Zhang et~al.(2022)Zhang, Roller, Goyal, Artetxe, Chen, Chen, Dewan,
  Diab, Li, Lin et~al.}]{zhang2022opt}
Susan Zhang, Stephen Roller, Naman Goyal, Mikel Artetxe, Moya Chen, Shuohui
  Chen, Christopher Dewan, Mona Diab, Xian Li, Xi~Victoria Lin, et~al. 2022.
\newblock Opt: Open pre-trained transformer language models.
\newblock \emph{arXiv preprint arXiv:2205.01068}.

\end{thebibliography}
\bibliographystyle{acl_natbib}

\appendix

\section{Appendix}
\label{sec:appendix}

\subsection{CoT prompts}
\begin{figure}[ht]
\centering
\includegraphics[width=0.5\textwidth]{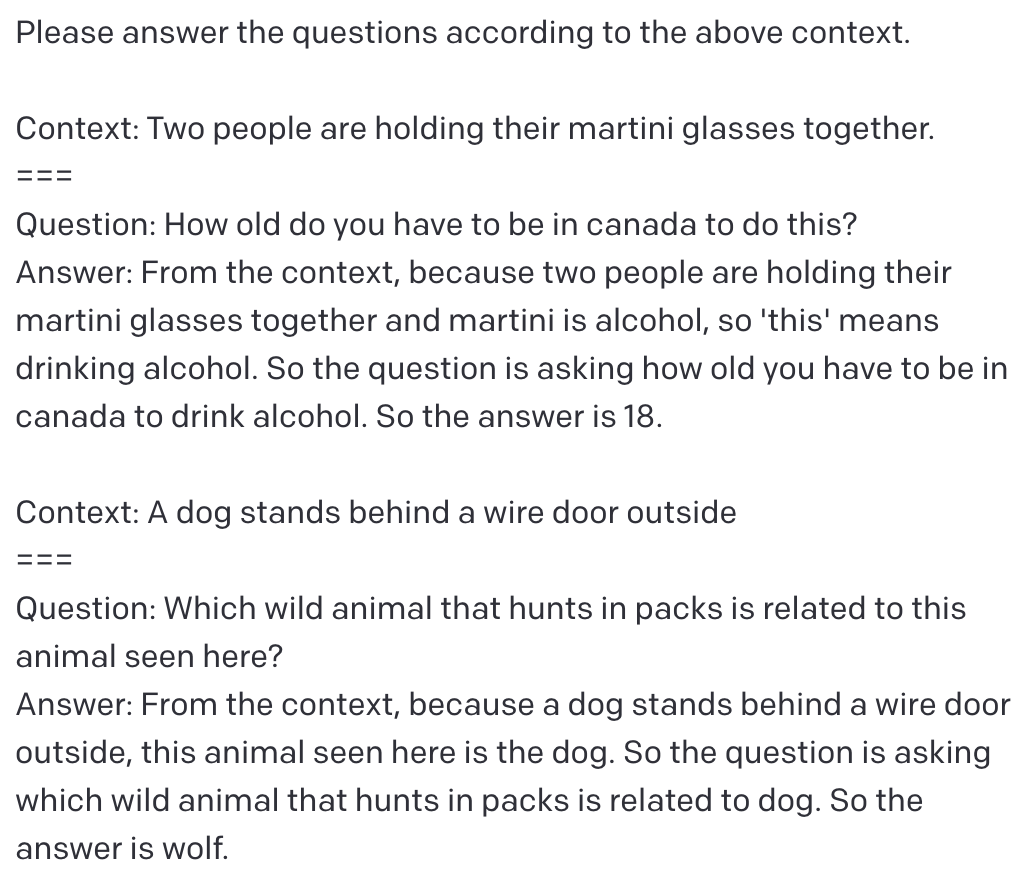}
\caption{The fixed prompt we use to generate chain-of-thought style rationales. We randomly select seven examples in the prompt and show two of them here. We set the temperature as 0.7 and max token as 80 during inference for all PLMs.}
\label{fig:cot}
\end{figure}

For our CoT generation experiments, we use a pre-designed fixed prompt as partly shown in Figure \ref{fig:cot}.

\subsection{Additional Experiments}
We conduct additional experiments for \modelname on an augmented successor dataset of OK-VQA: A-OKVQA \cite{AOKVQA} to prove its effectiveness.
Since we do not have the baseline results or any intermediate outputs on A-OKVQA as the paper was written, we only compare with PiCA \cite{yang2021empirical} with a simpler setting: without using image tagging or chain-of-thought and only using GPT-J.
The captions we use are generated using BLIP-2 \cite{li2023blip2}, following the default example in the paper.

\begin{table}[ht]
% \small
\centering
\setlength{\tabcolsep}{14pt}
\begin{tabular}{c|cc}
\hline
 & PiCA & \modelname\\ \hline
A-OKVQA  & 33.2  & \textbf{37.1}  \\
\bottomrule
\end{tabular}
\caption{Additional comparison of \modelname versus PiCA on A-OKVQA dataset.}
\label{tbl:app}
\vspace{0.5em}
\end{table}

% \subsection{Answer Choice Generation}
% \label{sec:app:gen}

% \subsection{Answer Selection}
% \label{sec:app:select}

\end{document}